\documentclass[letterpaper, 10 pt, conference]{ieeeconf}  
\usepackage{url}
\usepackage{cite}
\usepackage{graphics} 
\usepackage{epsfig} 
\usepackage{mathptmx} 
\usepackage{times} 
\usepackage{amsmath} 
\usepackage{amssymb}  

\usepackage{gensymb}

\usepackage{epstopdf}
\usepackage[]{units}
\usepackage{color}
\usepackage{verbatim}
\usepackage{psfrag}
\usepackage{textcomp}
\usepackage{graphicx}
\usepackage{caption}
\usepackage{subcaption}
\usepackage{algorithmic}
\usepackage[Pseudocode]{algorithm}
\IEEEoverridecommandlockouts                              

\title{\LARGE \bf
MOMA: Visual Mobile Marker Odometry
}

\author{Raul Acuna$^{1}$, Zaijuan Li$^{1}$ and Volker Willert$^{1}$
\thanks{*This work was sponsored by the German Academic Exchange Service (DAAD) and the Becas Chile doctoral scholarship.}
\thanks{$^{1}$All authors are within the Institute of Automatic Control and Mechatronics, TU Darmstadt, Germany.{\tt\small (racuna, zaijuan.li, vwillert)}{\tt\small @rmr.tu-darmstadt.de}}}

\begin{document}
\maketitle
\thispagestyle{empty}
\pagestyle{empty}

\begin{abstract}
In this paper, we present a cooperative odometry scheme based on the detection of mobile markers in line with
the idea of cooperative positioning for multiple robots \cite{kurazume1994}. 
To this end, we introduce a simple optimization scheme that 
realizes visual mobile marker odometry via accurate fixed marker-based camera positioning
and analyse the characteristics of errors inherent to the method compared to
classical fixed marker-based navigation and visual odometry. 
In addition, we provide a specific UAV-UGV configuration that allows for continuous 
movements of the UAV without doing stops and a minimal \textit{caterpillar}-like configuration that works with one UGV alone.  
Finally, we present a real-world implementation and evaluation for the proposed UAV-UGV configuration.
\end{abstract}

 

\section{INTRODUCTION}

Visual pose estimation and localization is a problem of interest to many fields from robotics to augmented reality and autonomous cars. 
Possible solutions are dependent on the camera(s) configuration available to the task (monocular, stereoscopic or multi-camera), 
as well as the amount of knowledge about the structure and geometry of the environment. 

If a mobile multi-robot system is available, cooperative localization firstly introduced by Kurazume et al. \cite{kurazume1994} drastically speed-up and improve the accuracy of the localization of each of the robots \cite{fox2000, mourikis2006}. 
The original idea is to use some robots as moving landmarks and others to detect them.
This allows a mobile robot-marker system to localize itself in an unstructured environment lacking enough features.
A bunch of different realizations \cite{Clift2015} and extensions to multiple-robot \textit{SLAM} (simultaneous localization and mapping) have been investigated \cite{saeedi2016}.

Visual pose estimation can be classified into two different categories: The first one, called marker-based (\textit{MA}), 
relies on some detectable visual landmarks like fiducial markers or 3D scene models
with known coordinates of its features/keypoints \cite{aruco,  Marchand2016}. 
The second category works markerless (\textit{MAL}) without any 3D scene knowledge \cite{Fraundorfer2012a, Marchand2016}.

\textit{MA} methods estimate the relative pose to a marker with known absolute coordinates in the scene. Therefore, these methods are driftless, 
need only a monocular camera system, and the accuracy of the pose estimation is both dependent on the accuracy of the measurement of 2D image coordinates of known 3D marker coordinates
and on what kind of algorithm is used to realize spatial resection \cite{willert2010, haendler2012}.

\textit{MAL} methods estimate relative poses between camera frames based on static scene features with unknown absolute coordinates in the scene and apply dead reckoning to reach the absolute pose within the scene in relation to a known initial pose. 
Due to this incremental estimation, errors are introduced and are accumulated by each new frame-to-frame motion estimation, which causes unavoidable drift. These methods can further be divided into pure visual odometry (\textit{VO}) \cite{Fraundorfer2012a} and more elaborate visual simultaneous localization and mapping (\textit{V-SLAM}) approaches \cite{lemaire2007} including the new developments on Semi-Dense visual odometry\cite{Engel2013b}. Basic \textit{VO} approaches estimate frame-to-frame pose changes of a camera based on some 2D feature coordinates, their optical flow estimates \cite{willert2009} and  their 3D reconstruction using epipolar geometry in conjunction with an outlier rejection scheme to verify static features \cite{Buczko2016}.
Even if some additional temporal filtering like extended Kalman filtering (\textit{EKF}) or local bundle adjustment (\textit{BA}) is applied, drift can be reduced but cannot be avoided \cite{Fraundorfer2012a}.

\textit{V-SLAM} approaches \cite{lemaire2007} not only accumulate camera poses but also 3D reconstructions of the back-projected extracted 2D features of \textit{VO} in a global 3D map. 
Thus, drift can be reduced using additional temporal filtering on the 3D coordinates of the features in the map or global \textit{BA} and \textit{loop closure} techniques 
to relocate already seen features via map matching.   
Both approaches can be realized with a monocular or a stereo vision system, whereas the stereo approach
is much less prone to drift because of the superior resolution of scale estimates. 
Alternatively, additional sensors like IMU can be integrated to improve the scale/drift problem in monocular systems and apply sensor fusion to increase robustness and reduce the drift as in Visual Inertial Odometry approaches~\cite{Leutenegger2015}. 

The main advantage of \textit{MA} versus \textit{MAL} methods (besides the fact that it does not drift) is the knowledge of error free 3D coordinates of easy and unambiguously detectable landmarks.
Thus, for \textit{MA} methods the error of spatial resection reduces to errors in 2D coordinate estimation of known 3D coordinates projected onto the image plane \cite{haendler2012}.
In contrast, \textit{MAL} methods have to deal with additional errors, 
like 1) outliers (e.g. non-static features), 2) 2D-2D correspondence errors from optical flow estimates and
3) 3D reconstruction errors stemming from inaccurate stereo vision, wrong disparities or scale estimations \cite{Buczko2016}. 

\textit{MAL} methods usually require good illumination (enough brightness and contrast) of the environment, scenes rich in texture and a certain amount of feature overlap between frames.

To summarize, each method has its own advantages and problems. In terms of accuracy and computational complexity \textit{MA} methods clearly 
outperform \textit{MAL} methods. The big advantage of \textit{MAL} methods is that only features which are already present in the environment
are needed for localization. Hence, it does not require the modification of the environment with artificial markers and/or a topological survey
to define landmarks covering the whole navigation space of the sensor.

The main motivation of our work is to develop a real-time cooperative visual localization method that keeps the accuracy of marker-based pose estimation 
without having the need to modify the environment. 
For this purpose, we propose a cooperative visual odometry scheme based on mobile visual markers (\textit{MOMA}). 

Our work is an extension of the Cooperative Positioning System method (CPS) based on mobile landmarks developed by Kurazume et al. \cite{kurazume1994}, 
but with a complete realization for the case when the landmarks are visual fiducial markers which can be detected with a monocular camera, e.g. Aruco markers \cite{aruco}. 
This avoids the need of using expensive laser based sensors. Additionally, a study of the propagation of the error was performed based on the particulars of monocular camera 
fiducial marker detection and its pros and cons compared to other popular feature based \textit{VO} and \textit{V-SLAM} approaches.



Fiducial markers have been used for relative pose estimation and tracking in the robotic community for quite some time, 
e.g. as beacons for UAV autonomous landing~\cite{Li2011} or as landmarks for the relative pose estimation of an UAV to a group of UGV's~\cite{Clift2015}. 
Common coordinate for multi-robot systems are also a topic of interest. Wildermuth et al. used a camera system mounted on top of a robot to calculate 
the relative position of each surrounding robot and their transformations in a common coordinate frame~\cite{Wildermuth2003}. 
More recently, Dhiman et al. developed a system of mutual localization which uses reciprocal observation of fiducials for relative localization without egomotion estimates or mutually observable world landmarks~\cite{Dhiman2013}. To the best of our knowledge, the idea of cooperative visual odometry based on mobile visual markers
has not been published.

The paper is structured as follows: In Sec.~\ref{SecII}, we introduce the basic principle of the \textit{MOMA} odometry scheme including an analysis of possible error sources compared to other pose estimation systems. In Sec.~\ref{SecIII}, we present different configurations of multi-robot-systems suitable to apply \textit{MOMA} odometry. 
In Sec.~\ref{SecIV} a real robotic experiment is shown along with a comparison with state-of-the-art methods followed by an evaluation. 
We demonstrate that \textit{MOMA} odometry is a reliable and accurate pose estimation method, especially when applied in multi-robot systems
and summarize its pros and cons in Sec.~\ref{SecV}.

\section{Mobile Marker based odometry}
\label{SecII}

We define the concept of a Mobile Marker (\textit{MOMA}) as a regular marker (fiducial or other kind of known feature) 
that has two possible configurable states at any given time: \textbf{Mobile}, if the marker is moving or permitted to move and \textbf{Static} otherwise. 
A \textit{MOMA} can either be moved by some entity or by itself. We define the \textit{observer} as the entity that performs the detection and pose estimation of the marker, 
in our case a camera. In order to do this pose estimation, the camera also needs to have one of these two states at a given time, \textbf{Mobile} or \textbf{Static} and also needs to use them in a certain way 
depending on the state of the marker.

The pose $\mathbf{G}$ in homogeneous representation\footnote{$\mathbf{G} = \begin{bmatrix} \mathbf{R} & \mathbf{T}\\ 0_{1x3}    & 1   \\ \end{bmatrix}$} 
is given by the 3D translation vector $\mathbf{T} \in \mathbb{R}^3$  and the rotation matrix  $\mathbf{R} \in \mathbb{R}^{3\times3}$.

\paragraph{Marker-based visual localization (\textit{MOMA})}

\begin{figure}[t!]
    \centering  
    \vspace{0.20cm}
    \begin{subfigure}[b]{0.23\textwidth}
        \includegraphics[width=\textwidth]{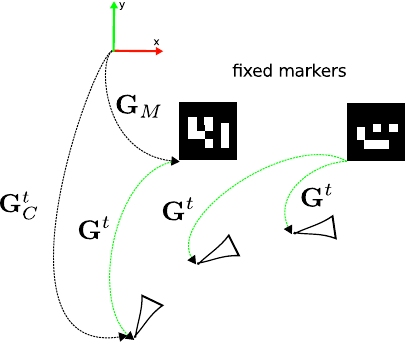}
        \caption{}
        \label{fig:marker_based_pose}
    \end{subfigure}
    ~ 
    \begin{subfigure}[b]{0.23\textwidth}
        \includegraphics[width=\textwidth]{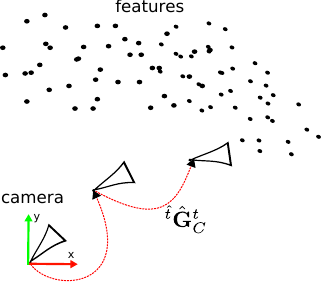}
        \caption{}
        \label{fig:visual_odometry}
    
    \end{subfigure}

    \caption{\small Camera-based pose estimation methods. Marker-based (\textit{MA}) pose estimation (a) uses known fixed markers with pose $\mathbf{G}_M$ 
             to obtain the absolute pose of the camera $\mathbf{G}_C^t$ at each time instant $t$ via the estimate $\mathbf{G}^t$ (in green).
             Visual odometry (b) detects fixed features along consecutive image frames in a markerless environment (\textit{MAL-VO}) to 
             estimate relative poses $(^{t}\mathbf{G}_C^{\hat{t}})$ (in red) and infers the absolute pose $\mathbf{G}_C^t$ by concatenation.}\label{fig:pose_estimation_methods}
\end{figure}

A marker is no more than a set of known features with known marker frame coordinates\footnote{All coordinates $\mathbf{X} = [X, Y, Z, 1]^T$ are assumed to be homogeneous coordinates, as long as not stated otherwise.} 
$\mathbf{X}_M$. Visual marker based pose estimation uses known fixed markers $M$ to obtain the absolute pose $\mathbf{G}_C^t$ of a camera $C$ at some time $t$ in 
world coordinates $\mathbf{X}_W$. We assume that the pose of the fixed marker in world coordinates $\mathbf{G}_M$ is known and also the structure of the marker is predefined and easy to detect. 
Once the marker is detected we can estimate the relative pose $\mathbf{G}^t$ of the marker in camera frame, and by extension the pose of the camera 

\begin{equation}
\mathbf{G}_C^t = \mathbf{G}^t \mathbf{G}_M
\end{equation}

in world coordinates using a PnP method. The error in global camera pose $\mathbf{G}_C^t$ will be only associated to the relative pose estimation between marker and camera $\mathbf{G}^t$. 
Hence, no drift will be accumulated as in dead reckoning approaches.

The reasons for the robustness and preciseness of a \textit{MA}-based pose estimate is twofold. First,
the 3D-2D correspondences $\{\mathbf{X}_M$, $\mathbf{x}^t\}$ can be extracted unambiguously using the knowledge about the configuration of the 3D points $\mathbf{X}_M$ on the marker \cite{aruco}.
Second, the coordinates $\mathbf{X}_M$ itself are known in advance from very precise measurements and do not have to be extracted online.
Thus, the only source for errors is the extraction of the coordinates of the 2D projections $\mathbf{x}^t$ which depends on the resolution of the camera
and the chosen method to get subpixel accuracy \cite{Marchand2016}. The relations for \textit{MA}-based pose estimations are sketched in Fig.~\ref{fig:marker_based_pose}.


\paragraph{Markerless visual odometry (MAL-VO)}

Contrary to marker based pose estimation visual odometry is a dead reckoning (coupled navigation) approach given some initial known pose $\mathbf{G}_C^{0}$. 
To get the absolute position of the camera $\mathbf{G}_C^{t}$ the relative frame poses between time $\tilde{t} = t-1$ and $t$, 
denoted $^{\tilde{t}}\mathbf{G}_C^{t}$, have to be estimated in order to get the absolute position via recursive accumulation:

\begin{equation}
\mathbf{G}_C^t = (^{t}\mathbf{G}_C^{\tilde{t}}) \mathbf{G}_C^{\tilde{t}} \,.
\label{eq6} 
\end{equation}

The relative pose can also be extracted from the following 3D-3D correspondence

\begin{equation}
\mathbf{X}_C^{\tilde{t}}  =  (^{\tilde{t}}\mathbf{G}_C^{t}) \mathbf{X}_C^{t}\,.
\end{equation}

Again including the collinearity equation, now the reprojection error between projected 3D coordinates $\mathbf{X}_C^{t}$ and 2D coordinates $\mathbf{x}^{\hat{t}}$ can be formulated
as follows:

\begin{equation}
\epsilon_2^t = |\!|\mathbf{x}^{\tilde{t}} - \pi ((^{\tilde{t}}\mathbf{G}_C^t) \mathbf{X}_C^t)|\!|_2\, .
\end{equation}

Solving the least squares optimization

\begin{equation}
^{\tilde{t}}\hat{\mathbf{G}}_C^t = \mbox{argmin}_{^{\tilde{t}}\mathbf{G}_C^{t}} \sum_{\mathbf{x}^{\tilde{t}},\mathbf{X}_C^t} \left(\epsilon_2^t\right)^2 \, , \label{Eq:ReproMinVisualOdom}
\end{equation}
leads to relative pose estimates $^{\tilde{t}}\hat{\mathbf{G}}_C^t$ (see also Fig.~\ref{fig:visual_odometry}).
The 3D coordinates $\mathbf{X}_C^t$ of the features are not known and their estimation change over time. 
Thus, they have to be reconstructed as $\mathbf{X}_C^t = \lambda^t \mathbf{x}^t$, for example using a stereo vision system that extracts 
the depth $\lambda^t$ of each 2D coordinate $\mathbf{x^t}$. Also a proper correspondence search to get the 2D-2D correspondences of 
$\{\mathbf{x}^{\tilde{t}}$, $\mathbf{x}^t\}$ coordinate pairs is needed for a proper reconstruction and a good optimization result from \eqref{Eq:ReproMinVisualOdom}. 
Unfortunately, a correspondence search in a \textit{MAL} environment is ambiguous and prone to errors 
because it is based on some optical flow algorithm \cite{willert2009}. 
Since this reconstruction is not error-free and accumulates along frames, the \textit{MAL-VO} pose estimation is worse than \textit{MA} pose estimation and prone to drift
because of equation \eqref{eq6}.

\paragraph{Mobile marker odometry (\textit{MOMA})}

\begin{figure*}[ht]
\centering
\vspace{0.20cm}
\includegraphics[width=.96\textwidth]{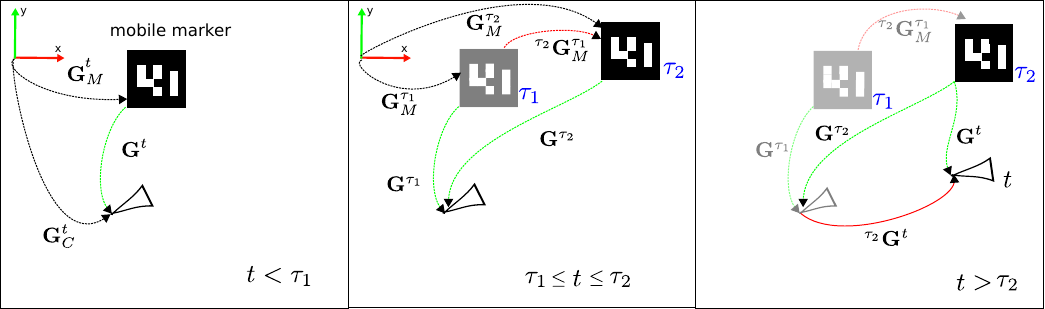}
\caption{\small The basic \textit{MOMA} odometry cycle. At $t=0$ the marker is static and the camera can obtain its initial pose $\mathbf{G}_C^0$ knowing the initial marker pose
$\mathbf{G}_M^{\tau_1}$. In timesteps $0 \leq t<\tau_1$ the camera moves in relation to the static marker and estimates its pose $\mathbf{G}_C^t$ by estimating the relative pose $\mathbf{G}^t$ to the marker.
During time $\tau_1 \leq t \leq \tau_2$ the camera is static and the marker starts to move to some new location in the FOV of the camera. Reaching time $t=\tau_2$
the marker stops moving and the marker pose change $^{\tau_2}\mathbf{G}_M^{\tau_1}$ can be estimated via $\mathbf{G}^{\tau_1}$ and $\mathbf{G}^{\tau_2}$.
Finally, starting from $t>\tau_2$ the marker is static again and the camera moves using the marker pose $\mathbf{G}_M^{\tau_2}$ as a new reference to estimate its pose $\mathbf{G}_C^t$, closing the cycle.}
\label{fig:moma_basic}
\vspace{-0.20cm}
\end{figure*}

In order to maintain the accuracy of fiducial marker pose estimation related to the camera $\mathbf{G}^t$ but using only one marker to cover the whole environment, 
the marker has to move. This means that the pose of the marker $\mathbf{G}_M^t$ may change at given time instances $t = \tau$ and the pose of the camera in world 
coordinates $\mathbf{G}_C^t$ is related to the marker pose via $\mathbf{G}^t$ as follows

\begin{equation}
\mathbf{G}_C^t = \mathbf{G}^t \mathbf{G}_M^{t=\tau}\, .
\label{eq:cam_pose}
\end{equation}

In order to get $\mathbf{G}_M^{t=\tau}$ at certain time instances $\tau$, the pose change $^{\tau_2}\mathbf{G}_M^{\tau_1}$ of the marker between two specific consecutive 
time instances $\tau_1, \tau_2$ with $\tau_2 > \tau_1$  has to be estimated.

Once this pose change is known, the current pose of the marker $\mathbf{G}_M^{\tau_2}$ can be recursively calculated from the last marker pose in $\tau_1$, which reads

\begin{equation}
\mathbf{G}_M^{\tau_2} = (^{\tau_2}\mathbf{G}_M^{\tau_1}) \mathbf{G}_M^{\tau_1}\, .
\label{eq:recursive_marker}
\end{equation}

Now we need to obtain this relative pose $^{\tau_2}\mathbf{G}_M^{\tau_1}$ by camera measurements. We start by fixing the camera into a static state with the following pose:

\begin{equation}
\mathbf{G}_C^{\tau_1} = \mathbf{G}^{\tau_1} \mathbf{G_M}^{\tau_1}\, .
\label{eq:tau1}
\end{equation}

For time interval $\tau_1 < t < \tau_2 $ the marker is in the mobile state and it moves to a new fixed pose in $\tau_2$ within the field of view (FOV) of the camera. 
Since the camera is static, the pose

\begin{equation}
\mathbf{G}_C^{\tau_2} = \mathbf{G}^{\tau_2} \mathbf{G_M}^{\tau_2}
\label{eq:tau2}
\end{equation}

is equal to $\mathbf{G}_C^{\tau_1}$. Hence, we can insert \eqref{eq:tau1} into \eqref{eq:tau2} and solve for the relative marker pose 

\begin{equation}
^{\tau_2}\mathbf{G}_M^{\tau_1} = [\mathbf{G}^{\tau_2}]^{-1} \mathbf{G}^{\tau_1}\, . 
\label{eq:tau3}
\end{equation}

The relative marker-camera poses $\mathbf{G}^{\tau_1}$ and $\mathbf{G}^{\tau_2}$ can be estimated and as long as the marker is static from time $\tau_2$ on, the camera can acquire its pose as in the fixed marker case for all times $t>\tau_2$.

Although there is drift by the accumulation of the relative poses of the marker according to \eqref{eq:recursive_marker}, 
as a matter of principle the accumulated error in \eqref{eq:recursive_marker} for mobile marker odometry is much lower than in \eqref{eq6} for visual odometry 
because no backprojection based on error-prone 3D reconstructions $\mathbf{X}_C^t$ has to be applied. 
Instead, only error-free marker coordinates $\mathbf{X}_M$ and very precise 3D-2D correspondences $\{\mathbf{X}_M$, $\mathbf{x}^t\}$ from a known fiducial marker that can be detected very robustly.  
Additionally, the error accumulation for \textit{MOMA} odometry according to \eqref{eq:recursive_marker} only happens at 
discrete time instances $t = \tau_i$ which occur on a much lower frequency at certain waypoints rather than on the frame rate of the camera like in \textit{MAL-VO}.

As a conclusion, the whole \textit{MOMA} odometry is only based on applying the least squares optimization along a specific \textit{caterpillar}-like 
(see also Sec.~\ref{SecIII}) marker-camera motion pattern. The minimal motion pattern and concurrent optimizations is summarized in a plain vanilla pseudocode \ref{alg} for visual \textit{MOMA} odometry. 

\begin{algorithm}
\begin{algorithmic} 
\STATE Initialize $\mathbf{G}_M^{\tau_1}$
\WHILE{$i$: marker localization cycles}
     \IF{$t=\tau_i$}
      \STATE marker and camera static: Detect marker to get $\mathbf{G}^{\tau_i}$
     \ELSIF{$\tau_i<t<\tau_{i+1}$}
      \STATE marker mobile and camera static: Continuously detect marker to get $\mathbf{G}^{t}$ and \eqref{eq:tau3}, \eqref{eq:recursive_marker} to get $\mathbf{G}_M^{t}$
     \ELSIF{$t=\tau_{i+1}$}
      \STATE marker and camera static: Detect marker to get $\mathbf{G}^{\tau_{i+1}}$
      and \eqref{eq:tau3}, \eqref{eq:recursive_marker} to get $\mathbf{G}_M^{\tau_{i+1}}$
     \ELSIF{$t>\tau_{i+1}$}
      \STATE marker static and camera mobile: Detect marker to get $\mathbf{G}^{t}$ and (\ref{eq:cam_pose}) to get $\mathbf{G}_C^t$
     \ENDIF
\ENDWHILE
\end{algorithmic}
\caption{Basic algorithm for visual \textit{MOMA} odometry}
\label{alg}
\end{algorithm}

The \textit{advantages} of the visual \textit{MOMA} odometry are: An \textbf{improved accuracy} with respect to other relative approaches like classical \textit{MAL-VO}. 
\textbf{Less computation time}, because the detection and pose estimation of e.g. Aruco fiducial markers takes around 10ms \cite{aruco} on a common 1 core PC, compared to realtime \textit{VO} for common 1 core PCs e.g. 30-60 ms \cite{Geiger2012CVPR}. In its basic configuration only a \textbf{monocular camera} is needed. 
An important advantage is that it doesn't require features in the environment and no intervention of the environment is needed
to setup the markers. Finally this method provides localization to the camera and the marker simultaneously even during movement. 
The \textit{disadvantages} are an increased control and navigation complexity and the need of communication or coordination 
between the marker and the camera since the marker now has an associated state.

The motion patterns for \textit{MOMA} odometry have the following movement restrictions:

\begin{enumerate}
\item The marker has to be static if the camera moves, and the camera has to be static as long as the marker moves. 
If more than one marker is used and one of the markers is static, then the camera is able to move all the time (which is not possible for \textit{CPS} \cite{kurazume1994}).
\item The marker and the camera move in turns.
\item During the transitions, static to moving or vice versa, there must be a period of time $dt$ where at least two devices are static (e.g. both camera and marker in a camera-marker configuration or
two markers in a camera-multi-marker configuration).
\end{enumerate}


A \textit{MOMA} implies new considerations in the classical robotics action-perception cycle. The action-perception cycle is based on the premise of act then perceive or perceive and then act. 
Now, in the \textit{MOMA} system we have what we call the perception-\textbf{interaction} cycle since the action of the marker affects the perception of the observer 
and in turn its action as well. 
The marker then can no longer be considered as a passive entity with no effect on the observer, a \textit{MOMA} is able to provide information regarding its current state to the observer, 
and the observer can also inform the \textit{MOMA} which state is needed for the general behaviour of the system in a given situation. 



\section{Possible Moma Robotic Architectures}
\label{SecIII}

In this section we will describe the possible robot configurations that we have considered based on monocular cameras and fiducial markers. In the experimental section the development and testing of a multi-robot system with one of these architectures will be shown.

\subsection{Caterpillar-like Configurations}

\begin{figure}
    \centering
    \vspace{0.20cm}
    \begin{subfigure}[b]{0.2\textwidth}
        \includegraphics[width=\textwidth]{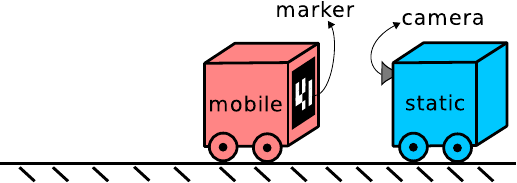}
        \caption{}
        \label{subfig:caterpillar_a}
    \end{subfigure}
    ~    
    \begin{subfigure}[b]{0.2\textwidth}
        \includegraphics[width=\textwidth]{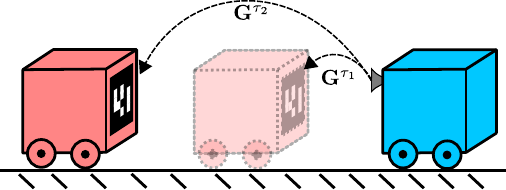}
        \caption{}
        \label{subfig:caterpillar_b}
    \end{subfigure}
    \linebreak
    
    \begin{subfigure}[b]{0.2\textwidth}
        \includegraphics[width=\textwidth]{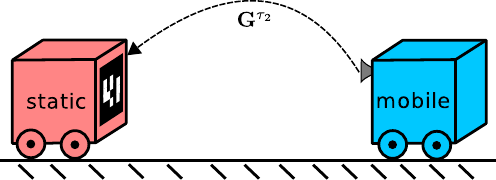}
        \caption{}
        \label{subfig:caterpillar_c}
    \end{subfigure}
    ~
    \begin{subfigure}[b]{0.2\textwidth}
        \includegraphics[width=\textwidth]{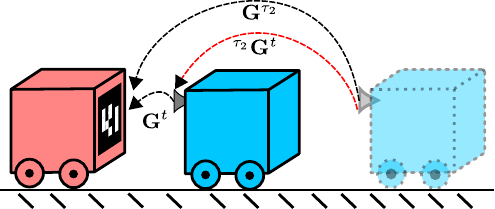}
        \caption{}
        \label{subfig:caterpillar_d}
    \end{subfigure}
    \linebreak
    \vspace{-0.5cm}
    \caption{Two-robot Caterpillar.}\label{fig:caterpillar}
    \vspace{0cm}
\end{figure}

This is the most basic multi-robot configuration for the \textit{MOMA} Odometry. It equals the structure we assumed in Sec.II to do the mathematical elaboration.

\subsubsection{Two-robot Caterpillar}

In this configuration one robot is the \textit{MOMA} (the one with the marker) and the other one is the \textit{observer} (the one with the camera), see Fig.~\ref{fig:caterpillar}. The \textit{observer} follows the movement of the \textit{MOMA} continuously thanks to the monocular camera. We named this particular kind of movement caterpillar-like motion, since each robot behaves like a segment of the body of a caterpillar. 


The \textit{MOMA} and the \textit{observer} move in turns, following the rules explained in Sec. \ref{SecII}. At the start, the \textit{observer} is static and the Moma is mobile and may move forward, Fig.~\ref{subfig:caterpillar_b}. Later on (Fig.~\ref{subfig:caterpillar_c}) the switching takes place, now \textit{MOMA} is static and the \textit{observer} is mobile. Finally, the \textit{observer} moves as in Fig.~\ref{subfig:caterpillar_d} and the  pose of the \textit{observer} is obtained from marker detection closing the cycle.


The error will be accumulated only during the switching of the reference and is only dependent on the accuracy of the fiducial marker detection, which by using a good camera and proper calibration may be in the range of millimetres~\cite{Bergamasco2013}. This system is also able to track the pose of the robots during the movement and not only in the transitions.

\subsubsection{Single-robot Caterpillar}

\begin{figure}
	\vspace{0.20cm}
    \centering
    \begin{subfigure}[b]{0.2\textwidth}
        \includegraphics[width=\textwidth]{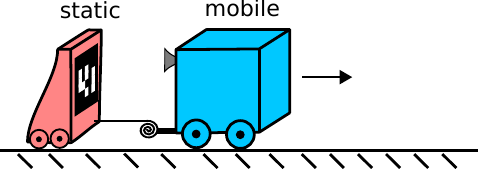}
        \caption{}
        \label{}
    \end{subfigure}
    ~    
    \begin{subfigure}[b]{0.2\textwidth}
        \includegraphics[width=\textwidth]{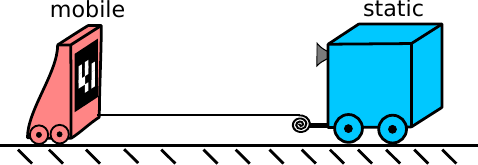}
        \caption{}
        \label{fig:tiger}
    \end{subfigure}
    \linebreak
    
    \begin{subfigure}[b]{0.2\textwidth}
        \includegraphics[width=\textwidth]{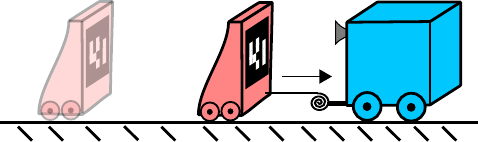}
        \caption{}
        \label{}
    \end{subfigure}
    \caption{Single-robot Caterpillar.}
    \label{fig:single_caterpillar}
    \vspace{0cm}
    \vspace{-0.50cm}
\end{figure}

In this minimal configuration only one robot will be pulling a sled with a simple pulley mechanism, see Fig.~\ref{fig:single_caterpillar}. The robot can either actuate to pull the sled close to himself or let it drag behind. A monocular camera detects a fiducial marker in the front of the sled. The robot performs caterpillar-like motion leaving the sled behind as static reference when it has to move, then stops and pulls the sled performing the \textit{MOMA} Odometry in the process.
  
\subsubsection{Multi-robot Caterpillar}

\begin{figure}[thpb]
	\centering
	\includegraphics[width=.3\textwidth]{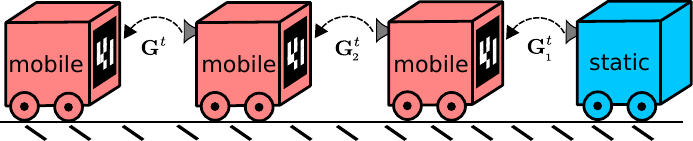}
    \caption{Multi-robot Caterpillar.}
    \label{fig:multi_caterpillar}
\end{figure}

This is an extension of the basic caterpillar case for $N$ robots, see Fig.~\ref{fig:multi_caterpillar}. Each robot follows the one in front. In this configuration $N-1$ robots with cameras are needed for the relative transformations. If at least one member of the group is static, the rest may move.

\subsection{Top Mobile Observer}

\begin{figure}[thpb]
	\vspace{-0.2cm}
	\centering
	\includegraphics[width=.25\textwidth]{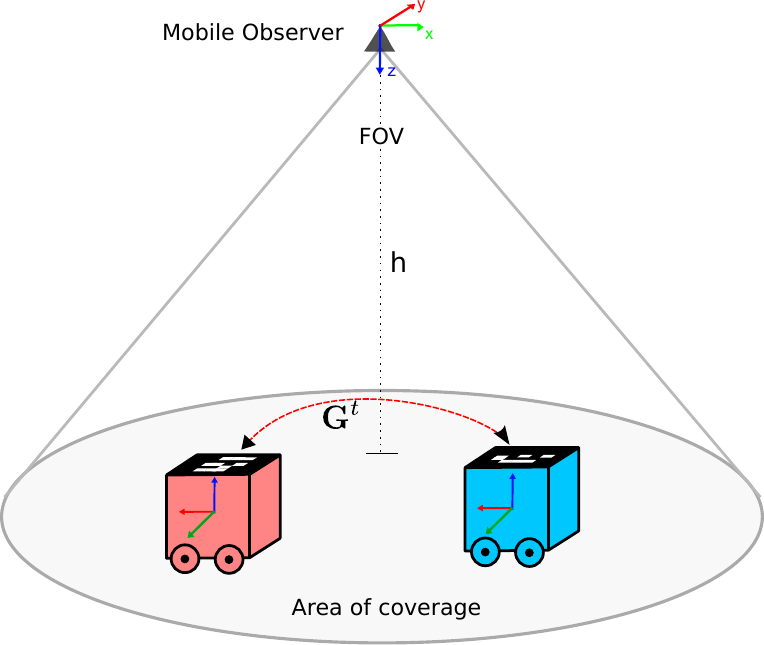}
    \caption{\small Top Mobile Observer.}
    \vspace{-0.2cm}
      \label{fig:top_observer}
\end{figure}

This configuration is based on two or more UGV's with fiducial markers on top and an external mobile \textit{observer} (UAV) which looks down to all the robots simultaneously using a monocular camera, see Fig~\ref{fig:top_observer}. The UGV's move in turns as in \textit{MOMA} Odometry but the \textit{observer} is  totally mobile.

The permitted action space on the ground $S = f(FOV, h)$ (area of coverage) for the movement of each robot will be a back projection of the field of view $\textit{FOV}$ of the camera on the ground plane dependent on the height from the camera to the ground $h$. Ideally, this coverage area will be centered in the middle of the UGV formation and the \textit{observer} should adjust its pose in order to cover the major amount of the image with all the markers. If the robots are close together the $h$ of the \textit{observer} should decrease to improve the marker detection, and if they move further apart the \textit{observer} has to move up in order to keep the markers inside the $\textit{FOV}$.

The \textit{observer} is a very general concept in this configuration, one logical choice is a quadcopter or any other type of UAV with a bottom camera. However, in our tests we also used a wireless camera in the hand of a person following the robots around the lab. An advantage of this configuration is that the \textit{MOMA} Odometry system will also fully locate the \textit{observer} and the \textit{observer} is always allowed to be in continuous movement. A further advantage of measuring the relative pose between markers from a top observer is that the resolution of the camera is exploited equally for each of the marker-camera pose estimates, because of the same distance from camera to markers. This contributes to more precise estimates compared to the situation where the marker-camera poses are at different distances and the camera resolution can't be optimally exploited.

\section{Experiments on a multi-robot system}
\label{SecIV}

The Top Mobile Observer configuration (Fig~\ref{fig:top_observer}) is more interesting because the area of coverage may be used as a local navigation space, with less robot movement restrictions than in the Caterpillar case. Hence, it was chosen to verify the accuracy of the Moma Odometry concept. This is also relevant in our group due to past research in the area of tracking and coverage using UAV's and UGV's~\cite{Khodaverdian}.

\subsection{Hardware configuration}

\begin{figure}[thpb]
	\centering
	\includegraphics[width=.25\textwidth]{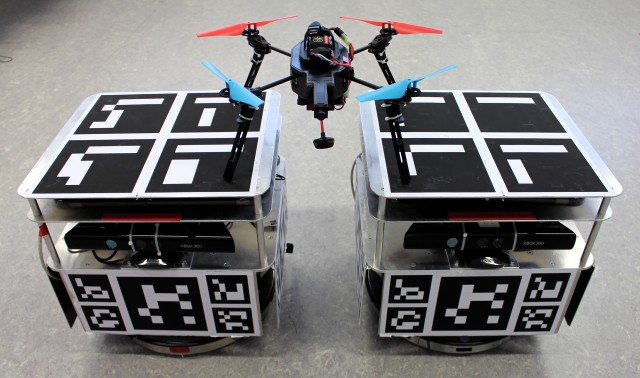}
    \caption{Robots used in our experiments.}
    \vspace{-0.2cm}
      \label{fig:robots_lab}
\end{figure}

Our experimental setup consists of two omnidirectional robots (Robotino\textsuperscript{\textregistered} ~from Festo Didactic Inc.). Each Robotino has an Aruco marker board on top, see Fig.~\ref{fig:robots_lab}. A wireless camera system was used for marker detection using a common configuration found in first person view racing drones. The video feed from the camera is transmitted to a ground station, digitized and processed by the PC (PAL format at $25 fps$). The UAV is an Ar.Drone 2.0 quadcopter with the wireless camera attached to the bottom and custom landing legs. 



\subsection{Software architecture}

\begin{figure}[thpb]
	\centering
	\includegraphics[width=.48\textwidth]{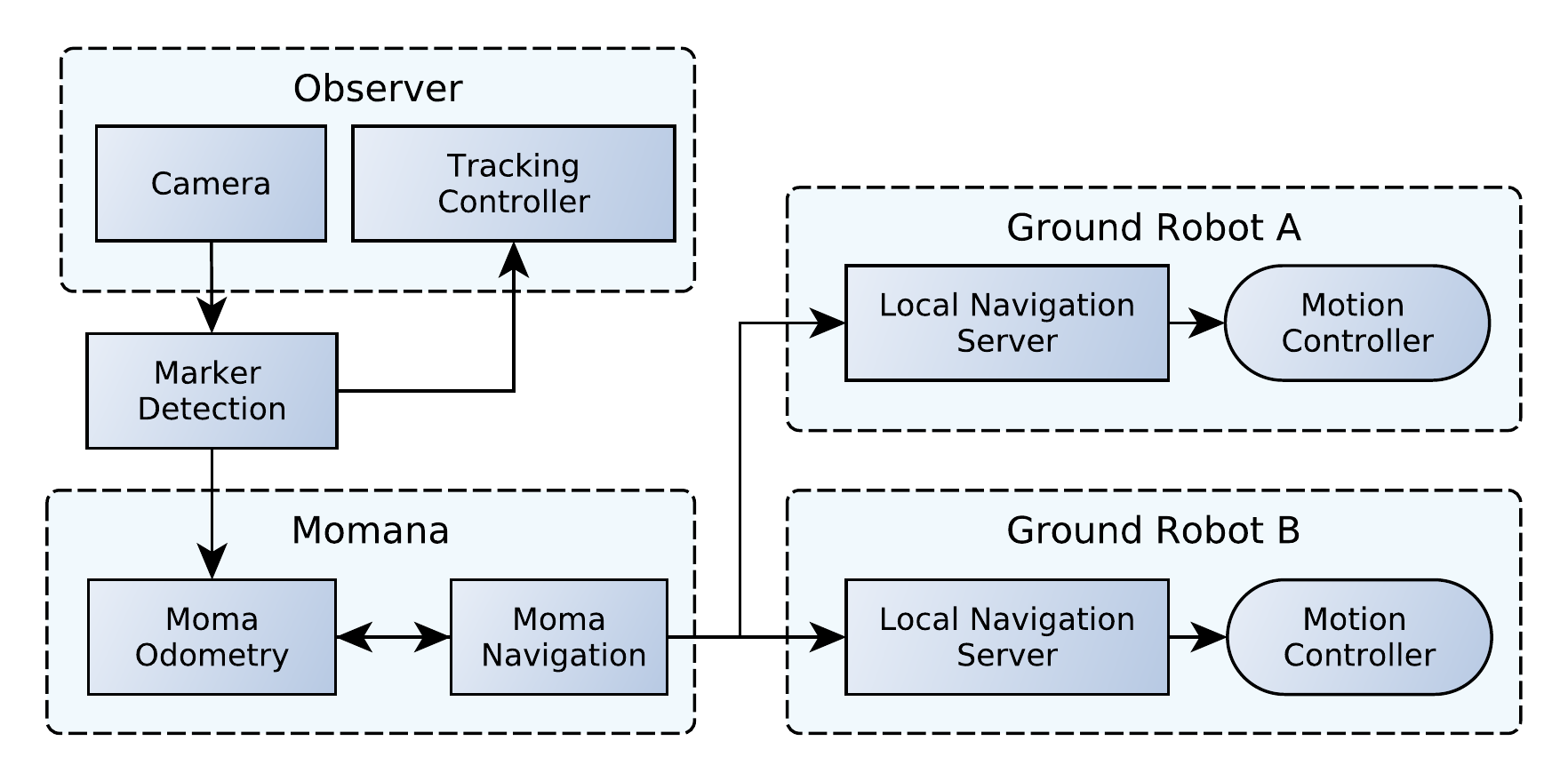}
    \caption{Moma Odometry and multi-robot navigation system.}
      \label{momana_block}
\end{figure}

The modules which are part of our system are shown in Fig.~\ref{momana_block}.  They comprise the tasks of marker detection, \textit{MOMA} odometry estimation, global navigation planner (\textit{MOMA} navigation), local navigation planner and motion control of the ground robots and the quadcopter. The code was implemented in the Robot Operating System (ROS) framework and is openly available at our research's group github account\footnote{\url{http://github.com/tud-rmr/tud_momana}}.

\subsubsection{Marker Detection module}

The flow of processing in the system starts at the \textit{observer} level, where the images from the camera are captured and sent to the PC for marker detection. We use the ROS package \textit{ar\_sys}, which is a wrap of the Aruco detection library with ROS functionality for the detection. Additionally, we coded a pre-processing ROS package (\textit{tud\_img\_prep}\footnote{\url{http://github.com/tud-rmr/tud_img_prep}}) in charge of de-interlacing and adjusting the input image for optimal marker detection. The final outputs are the poses of all the markers detected in the image in camera coordinate frame.

\subsubsection{Tracking Controller}

The control of the quadcopter for proper tracking of the ground robots using the detected marker poses is implemented in this module. Note that only relative poses are needed for tracking, even though the \textit{MOMA} Odometry system is able to provide it. The tracking control adjusts automatically the height, orientation and position of the quadcopter for optimal camera placement and marker detection.

\subsubsection{Moma Odometry}

Here, the algorithmic part of \textit{MOMA} Odometry is implemented as explained in Section \ref{SecII}. This module uses the detected markers in the camera frame to calculate the relative robot poses, it also tracks the current state of each \textit{MOMA} (mobile or static) and calculates the pose of all the robots in the system, including the \textit{observer} in the odometry coordinate frame.



There must be a time frame $dt$ where both robots need to be static during the switching. Since the \textit{observer} is mobile, the relative transform measurements will vary slightly with different \textit{observer} positions (due to camera calibration errors and image noise), so at any time the observer keeps an history of all the previous position estimates which then are used when both robots are static to calculate a better  $\mathbf{G}^{t}$ transform during the switch, which serves as a good and simple strategy to minimize the accumulation of error.

\subsubsection{Moma Navigation}

The ROS navigation stack is used to calculate the navigation path from a current UGV pose to the next goal from a series of predefined waypoints. The calculated paths are then sent to ROS local move servers running on the laptops of each Robotino, which are in charge of performing the path-following. The ROS navigation stack currently is not properly adapted to multi-robot configurations, this means that the goals for the robots need to be configured manually by the user, taking in consideration the movement constrain of the \textit{MOMA} Odometry scheme.

\subsection{Experimentation and discussion}

\subsubsection{Waypoint navigation}

A simple navigation task was defined for our robotic system as a set of goals that form a square shape (side=1m). Each goal is a position and orientation in the map coordinate frame $goal = (x,y,\theta)$.  The navigation between the goals was performed using Moma Odometry and Moma Navigation.

In this experiment we wanted to compare the behaviour of our system to a VO approach in an environment that does not provide enough features for the VO. The square shaped navigation was performed in our laboratory, which has white walls, a radiator with a repetitive pattern and a floor without texture. This lack of features is usually a problem for VO systems. We added patterns rich in texture for the first half of the trajectory in the field of vision of the camera, while the second half was left without modification. As ground truth we used fixed ceiling HD cameras(\textit{MA}) and we chose Viso2~\cite{Geiger2011IV} as the VO system. The final metric of comparison was defined as the final pose of the main robot after performing a loop measured by ceiling cameras. We calibrated the top $marker$ coordinate frame and the $camera$ of the robot offline using our marker-camera ROS calibration package~\footnote{\url{http://github.com/tud-rmr/tud_calibration}} and we calibrated the intrinsic parameters of the cameras using the standard ROS Calibration package.

\begin{figure}[thpb]
	\vspace{0.20cm}
	\centering
	\includegraphics[width=.45\textwidth]{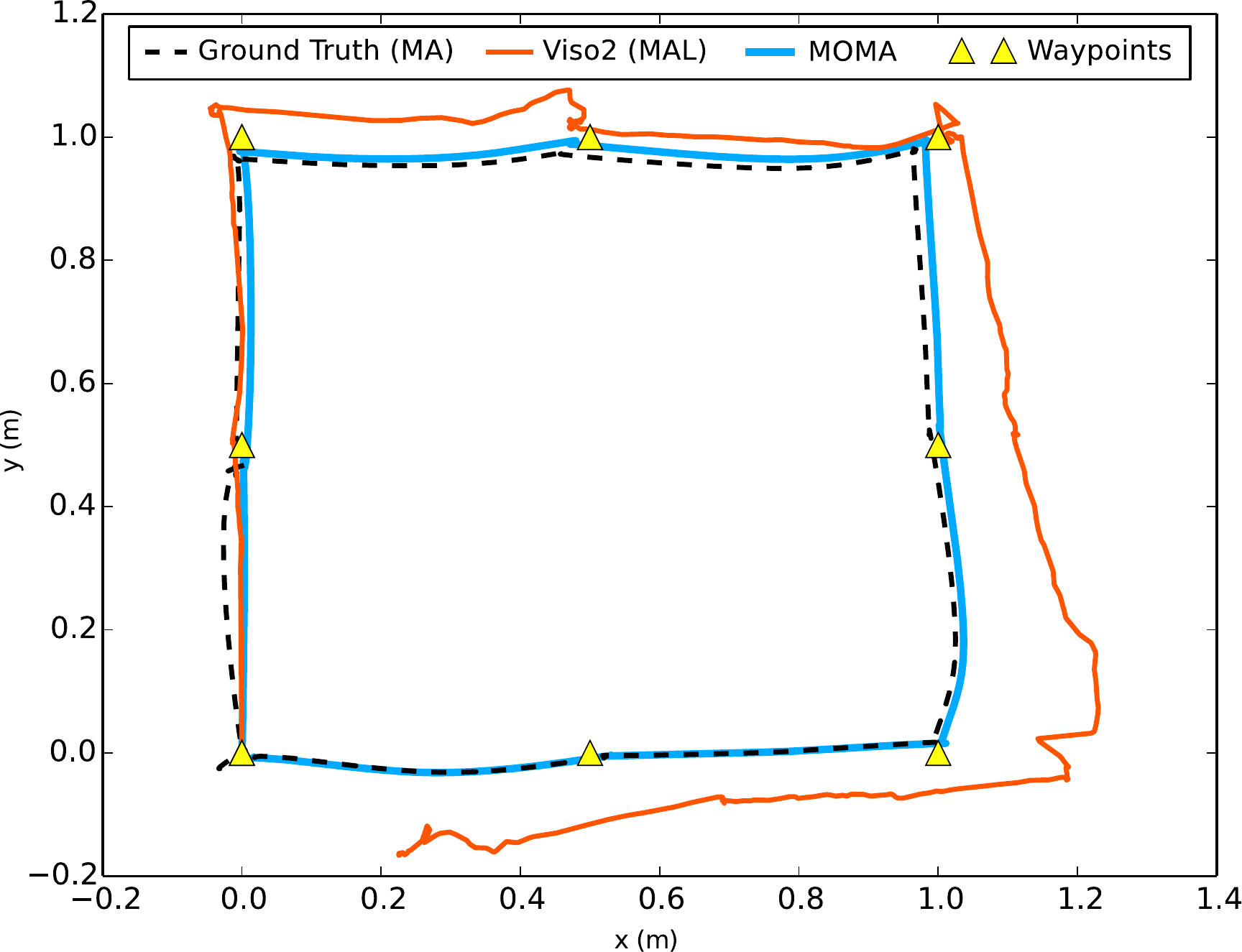}
    \caption{\small Odometry results for the main robot after waypoint navigation. In red is shown the behaviour of Viso2, which how the errors increases during rotations and due to the lack of good features in and indoor environment. Moma based odometry (blue) follows the waypoints with low error.}
    \vspace{-0.50cm}
    \label{fig:waypoint_result}
\end{figure}

In Fig.~\ref{fig:waypoint_result}, the result of one of the experiments is shown. For clarity, only the odometry information related to the main UGV is displayed. The blue solid line shows the odometry estimation using our proposed system (\textit{MOMA}), the black dashed is the ground truth (\textit{MA}) and the red solid line is the odometry estimation using visual odometry. The waypoints for the main robot are represented using yellow triangles. 

Our odometry system follows the trajectory measured by the ground truth with great accuracy and is able to easily track the trajectory of the robots at all times even between transitions. The odometry estimation of VO is also accurate as long as there are enough features in the environment (first half of the trajectory) and the movement does not include pure rotations. When the main robot performs pure rotations at waypoint coordinates $(0,1), (1,1)$ and $(1,0)$, the error in the pose estimation for the VO case increases sharply. This is an expected behaviour for \textit{MAL} based methods and confirms the advantages shown in Section~\ref{SecII} of the \textit{MOMA} odometry system. In our tests we found an additional serious problem related to VO in cooperative robotic systems. The movement of other robots disturbs the measurements, e.g. if a robot moves too close to the camera it may occlude good static features.

The waypoint navigation task was executed 10 times in our robotic system with different configurations, using the UAV as \textit{observer} and using a human with a hand-held camera as \textit{observer}. The error of the estimation is defined as the euclidean distance between the position obtained by a given method and the position given by the ground truth ($E$), we then calculated the mean error for the trajectory $ME$. Our main metric of comparison was the error of the final position ($E_f$) after performing the navigation task and the mean of the $E_f$ for all the tests was $ME_f$.

For our proposed method (\textit{MOMA}) the mean error on the final position was $ME_f = 0.97cm$ ($std=1.51$) which in percentage of the total trajectory (400cm) is  $0.2425\%$, with a $ME = 1.97cm$ ($std=0.69$). For Viso2 (\textit{MAL}) we obtained a $ME_f = 33.08cm $ ($std=16.42$), which in percentage of the total trajectory (400cm) is  $8.27\%$, with a $ME = 16.29cm$ ($std=6.98$),  this only includes the cases were Viso2 didn't lose track, which happened in almost 40\% of our tests. Our system's best case has $0.12\%$ of error for the total distance of the navigation task ($400cm$) while Viso2's best case was $1.32\%$.

\begin{figure}[thpb]
	\vspace{0.20cm}
    \centering
    \includegraphics[width=0.40\textwidth]{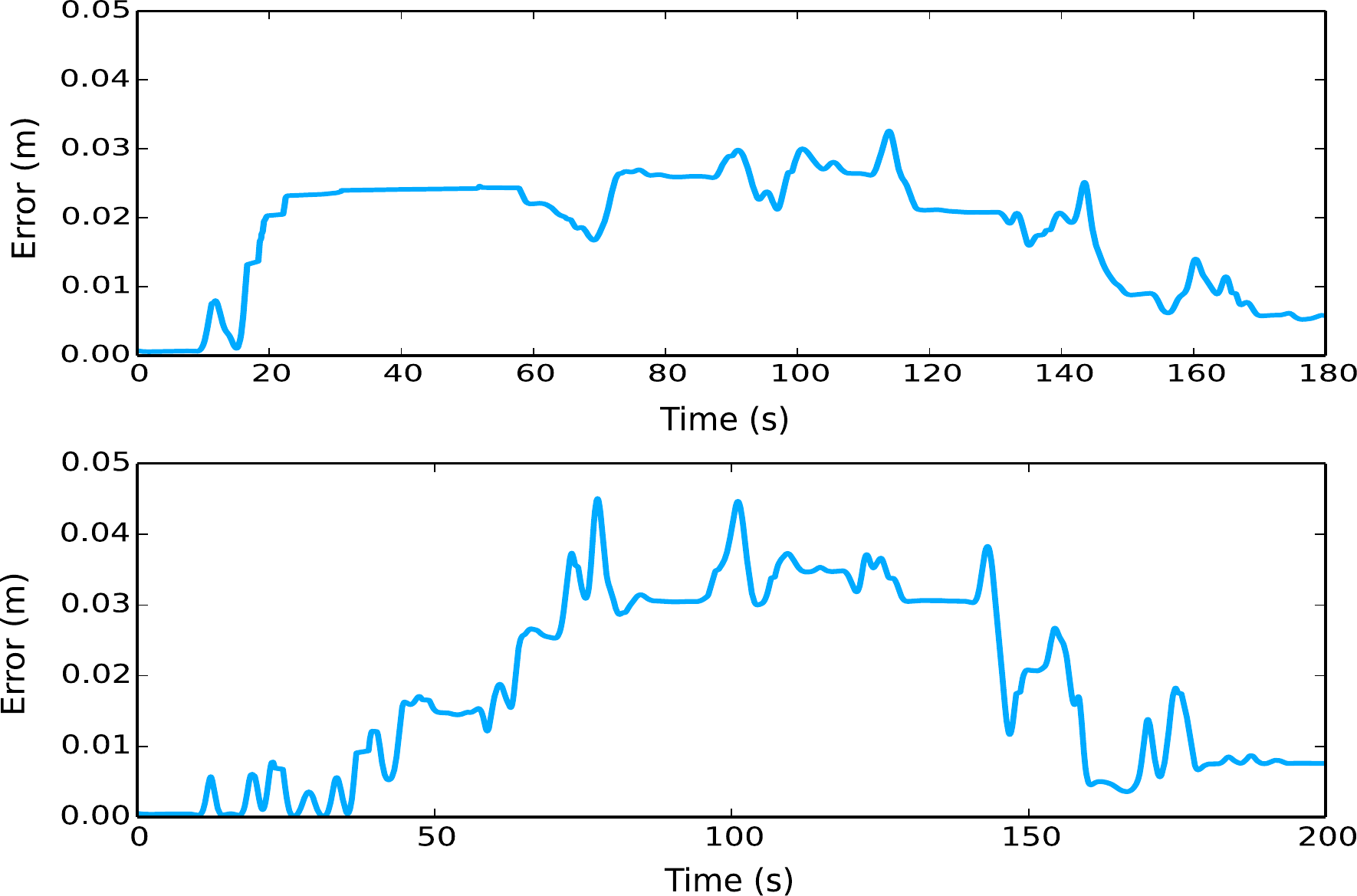}
    \caption{\small Comparison of the \textit{MOMA} odometry performance when using a handheld camera and a quadcopter.}
\label{fig:compare_human_quad}
\vspace{-0.50cm}
\end{figure}

In Fig.~\ref{fig:compare_human_quad} we perform a comparison of the behaviour of the error during the navigation task for two different cases: When the \textit{observer} is the UAV and when it is a handheld camera. The error when using a handheld camera was less erratic than with the quadcopter. The error in both cases at the beginning was close to zero and at the end of the navigation it was less than 0.01m since the ground truth camera was calibrated for the starting position and the middle part of the trajectory is in the border of the ceiling camera \textit{FOV}.

\subsubsection{Line Following}
\begin{figure}[thpb]
	\centering
	\vspace{0.10cm}
	\includegraphics[width=.45\textwidth]{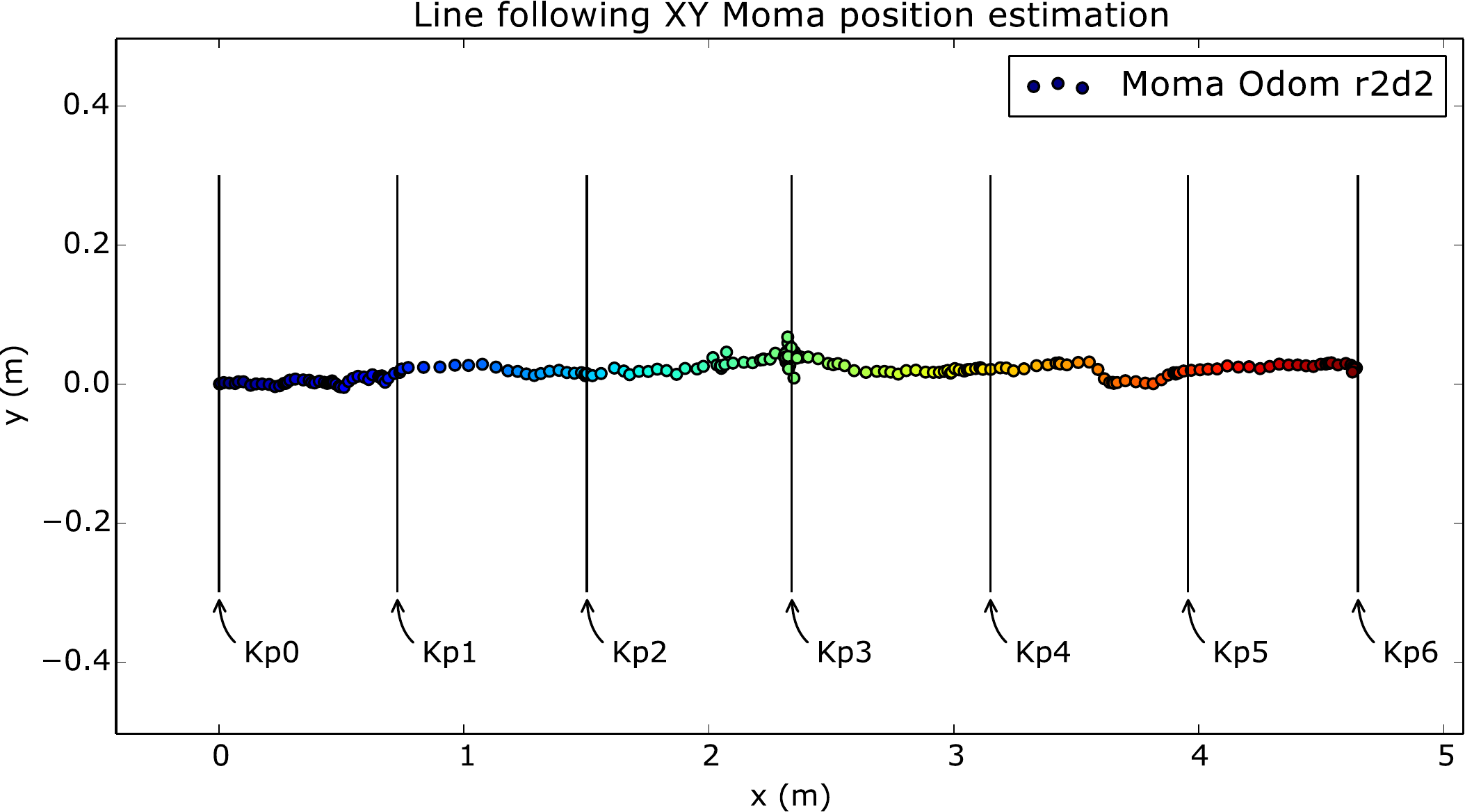}
    \caption{\small Odometry results for the main robot after one third of the line following navigation. The first six keypoints (switching instants) are represented by vertical lines.}
    \vspace{-0.20cm}
    \label{fig:line_following_result}
\end{figure}

\begin{figure}[thpb]
	\vspace{0.20cm}
	\centering
	\includegraphics[width=.45\textwidth]{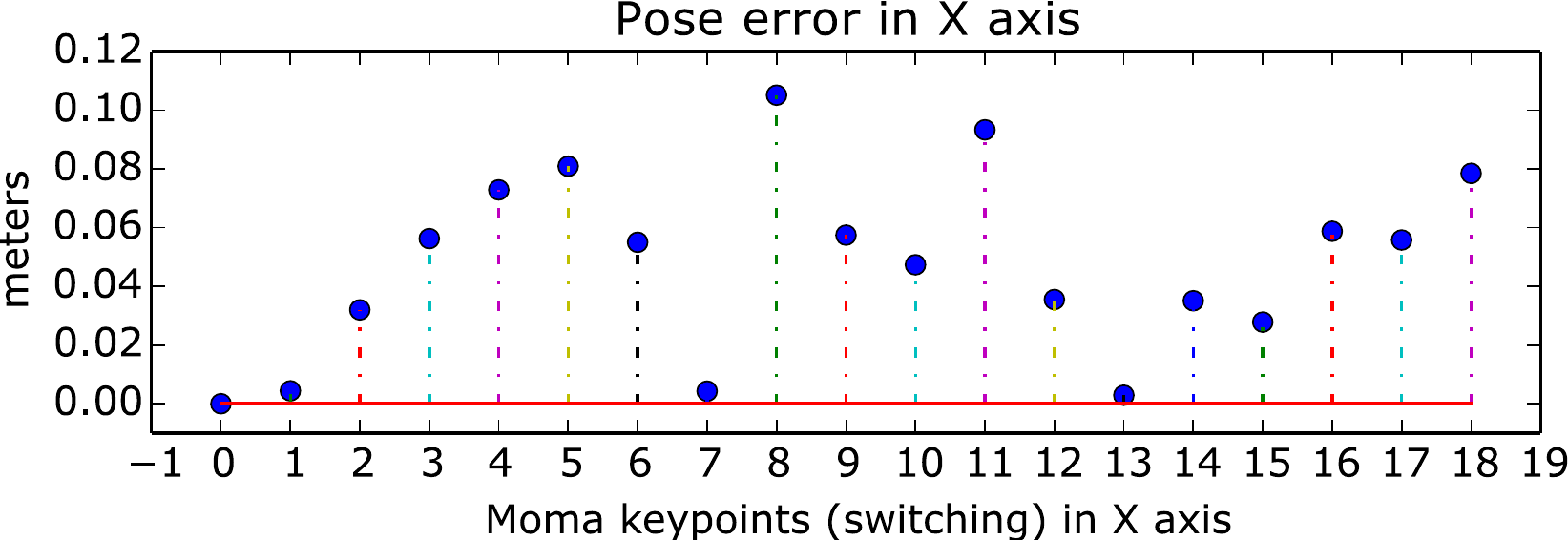}
    \caption{\small Pose estimate error for the line following test, the error is bounded and no greater that 0.1 meters. These are absolute values, in practice some errors cancel each other.}
    \vspace{-0.20cm}
    \label{fig:line_pose_error}
\end{figure}

\begin{figure}[thpb]
	\centering
	\vspace{0.00cm}
	\includegraphics[width=.45\textwidth]{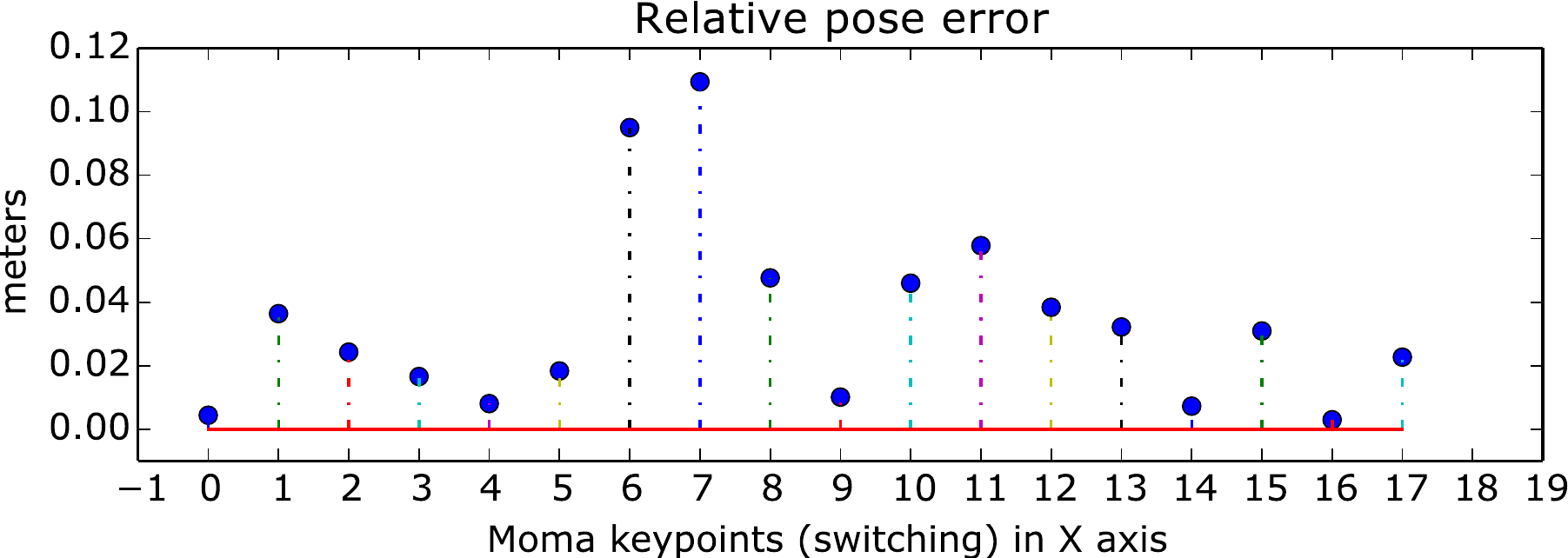}
    \caption{\small The absolute value of the error in relative measurements for the line following test.}
    \vspace{-0.40cm}
    \label{fig:line_relative_pose_error}
\end{figure}

This experiment was designed to show the behaviour of the error in Moma odometry. The main robot in a Top Observer configuration navigated a straight line of approximately 4.6 meters long three consecutive times (forward, backward and forward again) for an aproximate total distance of 13.785m. During the moments of the switching of reference (navigation keypoints) we measured with a laser the exact position of the robot in the X-axis. We used this as a ground truth to compare with the Moma estimation. During the course of a line segment navigation, 6 switching points (keypoints) were needed for a total of 18 for the whole trajectory. The results of the Moma estimation for one of the line segments and its corresponding keypoints are shown on Fig. \ref{fig:line_following_result}. The pose error in X axis, in each keypoint for the whole trajectory is shown in Fig. \ref{fig:line_pose_error}. We also show the absolute value of the relative pose error in Fig. \ref{fig:line_relative_pose_error}, that is the error in estimating the distance between keypoints. The final error after the 13.785m was 0.078m which correspond to 0.56\% of the total path. It is possible to observe that even though some of the errors are high (10cm) they get cancelled between each other giving the final performance.

As a final evaluation of MOMA, we believe that the proposed method could be an interesting tool for existing multirobot systems, since it provides a convenient solution for cooperative robotics in featureless environments. However, there is still great room for improvement. According to our simulations,  it is hard to obtain good relative measurements in the caterpillar-like motion (two UGVs), since the detection of Aruco markers does not provide good depth estimates (Z-axis of the camera). This may be solved by selecting other fiducial marker structures. The Top observer configuration is more precise since it is based on measurements on the XY plane of the camera, nonetheless, in order to give more freedom of movement for the UGVs, the UAV has to fly higher (decreasing marker detection accuracy) or the switching must happen when robots are in the border of the image (prone to distortion errors). Since the switching is the most critical part of the method (it is when the error accumulates), it is important to find new ways of improving the estimation accuracy by perhaps imposing additional constrains to the observer controller or by fusing the UAV's IMU measurements to counteract bad rotation estimates.

\section{Conclusions and future work}
\label{SecV}
We demonstrated a \textit{MOMA} Odometry system with greater accuracy than state-of-the-art MAL-based methods such as VO in featureless environments. Our proposed method is much easier to integrate into existing platforms since it only requires a cheap monocular camera and cheap fiducial markers, contrary to other methods. With our method no global positioning system like VICON is needed any more to conduct multi-robot navigation and control tasks. In future work we would like to improve measurement accuracy during transitions, e.g. by fusing the information from several robots observing each other and include the inertial sensors of the robots. Also, there is the necessity to implement a new layer (\textit{MOMA} Navigation) on top of the ROS navigation stack, where the user can define a goal for the system, or for any individual robot and \textit{MOMA} Navigation will calculate automatically the set of intermediate positions for each robot and execute the path-planning and path-following with the \textbf{•}\textit{MOMA} constraint.





{\small
\bibliography{IEEEabrv,momana_new}
}
\end{document}